\documentclass[letterpaper, 10 pt, conference]{ieeeconf}  
\usepackage[backend=bibtex]{biblatex}
\usepackage{amsmath}
\usepackage{tikz}
\usetikzlibrary{positioning, chains, shapes.geometric, fit, shapes, arrows.meta, calc, backgrounds}
\usepackage{booktabs}
\usepackage{multirow}
\usepackage{subcaption} 
\usepackage{cuted}
\usepackage{amssymb}
\IEEEoverridecommandlockouts    
\usepackage{censor}
\usepackage{comment}

\usepackage{float}
\usepackage{graphics} 

\graphicspath{ {./figures/} }

\title{\LARGE \bf LAR-MoE: Latent-Aligned Routing for Mixture of Experts in Robotic Imitation Learning}

\author{Ariel Rodriguez$^{1,2*}$, Chenpan Li$^{1*}$, Lorenzo Mazza$^{1}$, Rayan Younis$^{2,3}$, Ortrun Hellig$^{2,3}$, \\ Sebastian Bodenstedt$^{1,2}$, Martin Wagner$^{2,3}$, Stefanie Speidel$^{1,2}$
\thanks{*These authors contributed equally to this work.}
\thanks{$^{1}$Department of Translational Surgical Oncology, NCT/UCC Dresden, a
partnership between DKFZ, MF UKD, TUD, HZDR Dresden, Germany.}%
\thanks{$^{2}$Cluster of Excellence-CeTI, TUD, Germany.}
\thanks{$^{3}$Department of Visceral, Thoracic and Vascular Surgery, Faculty of
Medicine and University Hospital Carl Gustav Carus, TUD, Germany.}
}

\begin{document}

\maketitle

\thispagestyle{empty}
\pagestyle{empty}


\begin{strip}
    \centering
    \includegraphics[width=\textwidth,  trim={3cm 12cm 3cm 0cm}, clip]{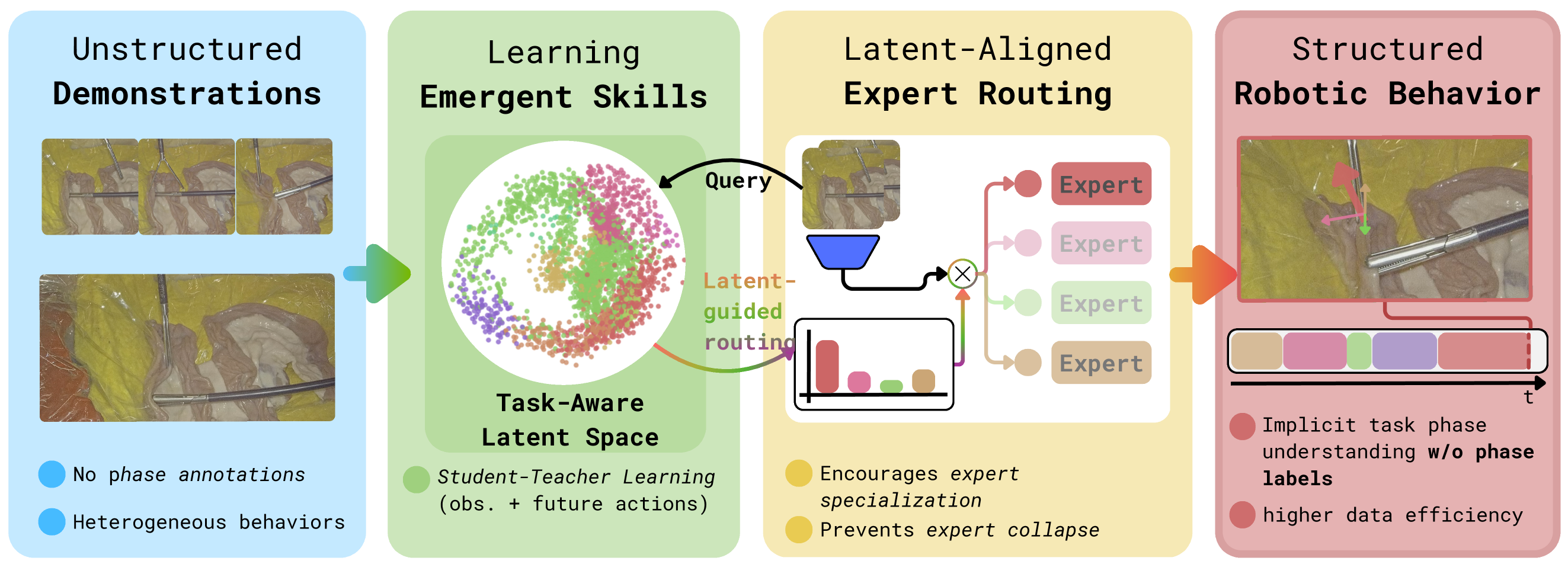}
    \captionof{figure}{Overview of the proposed LAR-MoE framework.
Unstructured demonstrations are first used to learn a task-aware latent space. The learned latent structure then guides expert routing, encouraging specialization while preventing expert collapse. This latent-aligned routing enables the emergence of structured robotic behavior through \textit{implicit task-phase understanding} without requiring explicit task phase annotations.}
    \label{fig_frontcover}
    \vspace{1em}
\end{strip}

\begin{abstract}

Imitation learning enables robots to acquire manipulation skills from demonstrations, yet deploying a policy across tasks with heterogeneous dynamics remains challenging, as models tend to average over distinct behavioral modes present in the demonstrations. Mixture-of-Experts (MoE) architectures address this by activating specialized subnetworks, but requires meaningful skill decompositions for expert routing. We introduce Latent-Aligned Routing for Mixture of Experts (LAR-MoE), a two-stage framework that decouples unsupervised skill discovery from policy learning. In pre-training, we learn a joint latent representation between observations and future actions through student–teacher co-training. In a post-training stage, the expert routing is regularized to follow the structure of the learned latent space, preventing expert collapse while maintaining parameter efficiency. We evaluate LAR-MoE in simulation and on hardware. On the LIBERO benchmark, our method achieves a 95.2\% average success rate with 150M parameters. On a surgical bowel grasping and retraction task, LAR-MoE matches a supervised MoE baseline without requiring any phase annotations, and transfers zero-shot to \textit{ex vivo} porcine tissue. Our findings suggest that latent-aligned routing provides a principled alternative to supervised skill decomposition, enabling structured expert specialization from unlabeled demonstrations.

\end{abstract}

\section{INTRODUCTION}
Imitation Learning (IL) has emerged as a promising paradigm by enabling robots to acquire complex skills without manual reward engineering \cite{brohan2023rt2, chi2024diffusionpolicy, zhao2023act}. More recently, the integration of large-scale vision--language foundation models with action data has led to the development of Vision--Language--Action (VLA) models, which demonstrate substantial improvements in open-world generalization capabilities \cite{intelligence2025pi05visionlanguageactionmodelopenworld}.

However, deploying a single policy across tasks with heterogeneous dynamics and contact patterns as seen in surgical tasks remains difficult, as these methods tend to average over distinct behavioral modes rather than specializing \cite{mandlekar2021matters}.
While recent progress has focused on scaling high-level models with increasing data availability, such approaches assume large, densely labeled datasets. In data-constrained domains like surgical robotics, demonstrations are scarce, requiring policies to learn from limited demonstrations.


Mixture of Experts (MoE) models offer a compelling solution by conditionally activating specialized subnetworks based on input characteristics \cite{fedus2022switch, riquelme2021scaling}. This sparse computation paradigm naturally aligns with robot
learning, where different manipulation primitives, such as reaching, grasping, or insertion, can be encoded by dedicated expert modules and dynamically selected based on visual context. While MoE architectures have achieved remarkable success in language and vision domains \cite{jiang2024mixtral, lin2024moe}, their application to visuomotor policy learning remains underexplored. A key challenge lies in discovering meaningful skill decompositions without explicit supervision or manual primitive definitions \cite{pertsch2021skild, shi2023skill} which is needed for expert routing.

To address this, we introduce \textbf{Latent-Aligned Routing for Mixture of Experts (LAR-MoE)}, a methodology that decouples representation learning from expert routing. We first learn a descriptive joint latent space of observations and future motions using an unsupervised co-training strategy inspired by student–teacher learning. This latent space captures the underlying task structure without explicit phase supervision. In a second stage, multiple compact action experts are combined through a soft routing mechanism \cite{obandoceron2024mixturesexpertsunlockparameter}, where expert selection is explicitly regularized to align with the structure of the learned latent space. By anchoring routing decisions to latent task similarities, LAR-MoE encourages specialization while mitigating expert collapse and maintaining parameter efficiency.

We evaluate our method on the LIBERO~\cite{liu2023libero} benchmark, demonstrating competitive performance while using significantly fewer policy parameters. Furthermore, we validate our approach on-hardware performing surgical task bowel grasping and retraction \cite{mazza2026moeact}.

Our main contributions are the following:
\begin{itemize} 
\item We introduce an unsupervised co-training strategy, to learn a descriptive joint latent space that captures the relationship between visual observations and future motion trajectories.
\item We propose the LAR-MoE architecture with a latent-alignment regularization strategy that anchors soft expert routing to the structure of the learned latent space, preventing expert collapse and significantly increasing parameter efficiency.
\item We show that routing structure can be learned from observation–future motion alignment without supervision by validating on the LIBERO simulation benchmark as well as a on-hardware surgical bowel grasping and retraction task performed on phantom and tested zero-shot on \textit{ex vivo} porcine tissue.
\end{itemize}

\section{METHODS}



\begin{figure*}[t]
    \centering
    \includegraphics[width=\textwidth, trim={0.5cm 10.5cm 0cm 1cm}, clip]{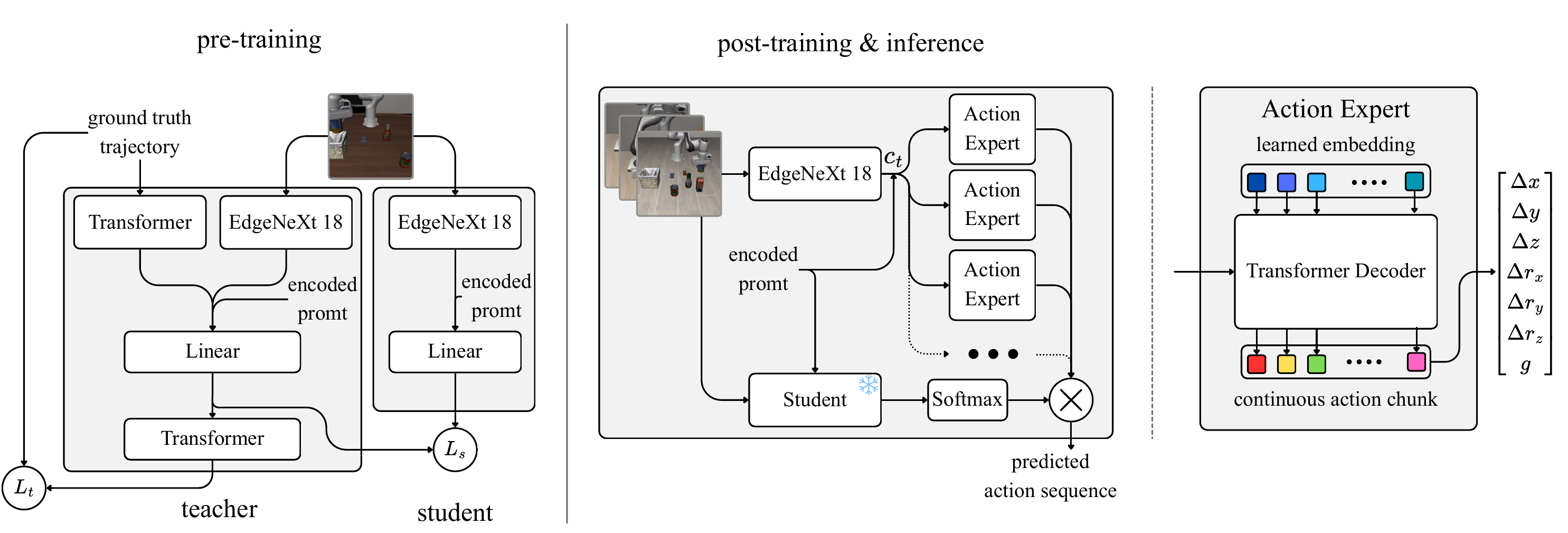}
    \caption{Overview of LAR-MoE. In pre-training, our method learns a joint latent representation of observations and future actions via student–teacher co-training. In post-training, we anchors expert routing to the learned latent structure by freezing the student model and using a regularization strategy. The action experts are implemented using a simple transformer decoder architecture. The language promts are encoded using MiniLM-L6.}
    \label{fig:model_diagram}
\end{figure*}

\subsection{Preliminaries}
Given a set of expert demonstrations $\mathcal{D}$, the goal of imitation learning is to maximize the log likelihood of the correct sequence of actions (also called action chunk of length $H$) $a_{t, t+H}$ selected under the current observation $o_t$ and a natural language task description $l$.

At each time step $t$, the observation $o_t$ consists of a set of multi-view images $o_t = \{\mathcal{I}_t^{1}, \mathcal{I}_t^{2}, \dots, \mathcal{I}_t^{n}\}$ capturing the scene from complementary viewpoints. The corresponding action $\mathbf{a}_t \in \mathbb{R}^7$ represents the change in the end-effector configuration and is defined as $\mathbf{a}_t = (\Delta \mathbf{x}_t, \Delta \mathbf{r}_t, g_t)$ where $\Delta \mathbf{x}_t \in \mathbb{R}^3$ denotes translation,  $\Delta \mathbf{r}_t \in \mathbb{R}^3$ represents rotational change, and $g_t \in \mathbb{R}$ specifies the gripper state.

\subsection{Policy Architecture}
Building on prior observations that complex manipulation tasks decompose into recurring subtasks \cite{li2017infogail, choi2025mixtureofactionexperts, obandoceron2024mixturesexpertsunlockparameter}, we model human demonstrations as arising from a mixture of latent policies, each capturing a distinct behavioral mode. We instantiate this hypothesis through a mixture-of-experts (MoE) architecture in which each expert specializes in a subset of the overall behavior, while a routing network maps observations to a discriminative task representation that correlates with the action distribution. Unlike standard end-to-end MoE training, which often suffers from expert imbalance—where frequently selected experts receive more gradient updates, further increasing their selection probability and leading to overfitting and underutilization—we decouple routing and policy learning. Specifically, we first pre-train the routing network to produce structured representations of the task space, and subsequently train the experts conditioned on these tasks.

\subsection{Router Network}
In the pre-training stage, we learn a joint latent representation of observations and actions via a co-training strategy inspired by student–teacher learning. Specifically, we optimize a student network $\phi_{s,\theta}$ to infer a latent vector $\hat{z}_t=\phi_{s,\theta}(o_t)$ from the current observation and match the teacher latent $z_t=\phi_{t,\theta}(o_t, a_{t:t+H})$ by minimizing $\mathcal{L}_{s} = \operatorname{MSE}(\hat{z}_t, z_t)$.
The teacher is optimized to reconstruct the action chunk by minimizing $\mathcal{L}_{t} = \operatorname{MSE}(\hat{a}_{t:t+H}, a_{t:t+H})$ with $\hat{a}_{t:t+H}=\psi(z_t)$ where $\psi(\cdot)$ is a trainable decoder.
This unsupervised pre-training allows the student network to learn the correlation between observations and action latent representation, whose structure can later be used for expert selection.

\subsection{Mixture-of-Experts Policy}
Our policy, shown in Fig. \ref{fig:model_diagram}, consists of a vision and language encoder, followed by $N$ action experts $(\psi_1, \psi_2, ..., \psi_N)$. Following the simple architecture of Action Chunking with Transformers (ACT) \cite{zhao2023act}, our action expert consists of a transformer decoder with learned embedding tokens to predict a chunk of actions at one step.

At post-training and inference, we use the frozen student model from pre-training to predict the action latent $\hat{z}_t$. Inspired by the stabilizing effect of soft gating in reinforcement learning \cite{obandoceron2024mixturesexpertsunlockparameter}, we employ a soft-gating mechanism to maintain gradient flow across experts and improve training stability. The learnable temperature parameter $T$ is initialized to 100.
\begin{equation}
    p_t = \text{softmax}\!\left( T \cdot \text{MLP}(\hat{z}_t) \right)
\end{equation}

Each expert receives a context vector $c_t$, formed by concatenating visual encodings from EdgeNeXt18~\cite{maaz2022edgenext} and representations from a frozen MiniLM-L6 language encoder \cite{wang2020minilmdeepselfattentiondistillation}, and produces an action chunk $\psi_n(c_t) = \hat{a}_{t:t+H,n}$, where $n$ indexes the expert. Using a weighted average, the final action chunk is computed.

\begin{equation}
    \hat{a}_{t:t+H} = \sum_{n=1}^{N} (p_t)_n \, (\hat{a}_{t:t+H})_n
\end{equation}

\subsection{Loss \& Regularization}
During post-training, we optimize the MSE loss $\mathcal{L}_{MSE} = \operatorname{MSE}(\hat{a}_{t:t+H}, a_{t:t+H})$ between our predicted action chunk and the demonstrated action chunk using the AdamW \cite{loshchilov2019decoupledweightdecayregularization} optimizer.

We propose the \textbf{distance consistency loss} that encourages the expert selection distribution to follow the distances of the task latents predicted by the student model. In this way, we mitigate expert balancing issues by learning a \textit{soft} task assignment for each expert.
\begin{equation}
    D^{(Z)}_{ij} = 1 - \frac{\mathbf{\hat{Z}}_i^\top \mathbf{\hat{Z}}_j}{\|\mathbf{\hat{Z}}_i\|_2 \, \|\mathbf{\hat{Z}}_j\|_2}
\end{equation}
\begin{equation}
    D^{(P)}_{ij} = 1 - \frac{\mathbf{\hat{P}}_i^\top \mathbf{\hat{P}}_j}{\|\mathbf{\hat{P}}_i\|_2 \, \|\mathbf{\hat{P}}_j\|_2}
\end{equation}
\begin{equation}
    \mathcal{L}_{\text{DC}}
    = \frac{1}{B^2}
    \left\|
    \mathbf{D}^{(Z)} - \mathbf{D}^{(P)}
    \right\|_F^2
\end{equation}
Distance consistency loss is computed over a batch, therefore $B$ is the batch size, $\mathbf{\hat{P}}=\{(p_t)_i\}_{i=1}^B$ and $\mathbf{\hat{Z}}=\{(\hat{z_t})_i\}_{i=1}^B$.

Further, we employ an \textbf{entropy regularization} to encourage experts to specialize to a certain task.
\begin{equation}
    \mathcal{L}_H = -\sum_{n=1}^{N} (p_t)_n \log (p_t)_n
\end{equation}

Following \cite{kang2025mixturegroupexpertslearning} we use \textbf{group sparse regularization} to further improve training stability as it has been shown to be effective on MoEs for image classification.

\begin{equation}
    \mathcal{L}_G = \sum_{i,j} \sqrt{\left( F_{\sigma} * \left( reshape(p_t)^2 \right) \right)_{ij}}
\end{equation}

We use the formulation exactly as defined in \cite{kang2025mixturegroupexpertslearning}, where $F_\sigma$ is a Gaussian lowpass filter with standard deviation $\sigma$, and $reshape(\cdot)$ reorganizes the routing probability vector $p_t$ into an approximately square matrix to enable spatial grouping of neighboring experts.

Our final loss function is a combination of the loss terms:

\begin{equation}
    \mathcal{L} = \mathcal{L}_{MSE} + \lambda_{\text{DC}}\,\mathcal{L}_{\text{DC}} + \lambda_H\,\mathcal{L}_H + \lambda_G\,\mathcal{L}_G
\end{equation}

The coefficients $\lambda_{\text{DC}}$, $\lambda_H$, and $\lambda_G$ control the relative contribution of each regularization term to the total loss.


\section{EXPERIMENTS AND EVALUATION}



%

\begin{figure*}[h]
    \centering
    \includegraphics[width=\textwidth, trim={0cm 0cm 0cm 0cm}, clip]{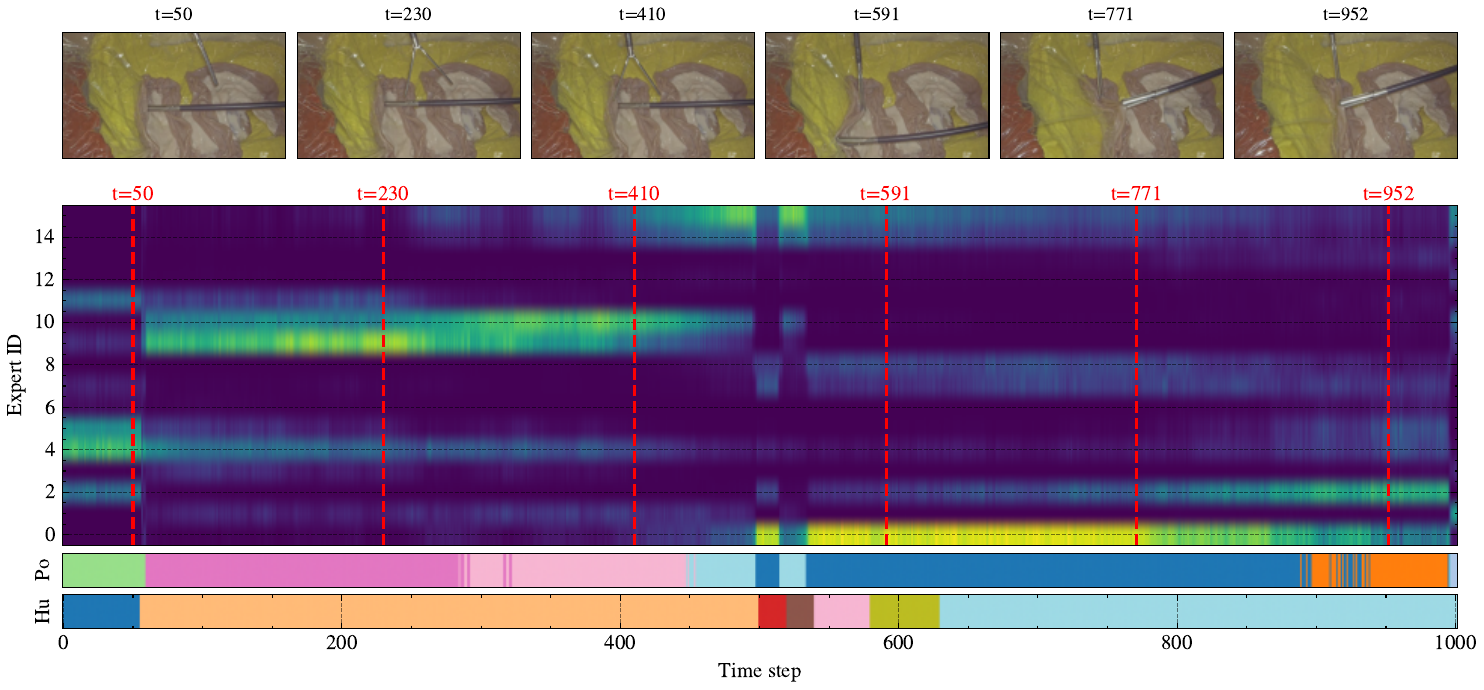}
    \caption{Expert activation over time during an ex-vivo bowel retraction rollout. The heatmap displays the activation weights of all experts at each timestep ($\approx 33\,\mathrm{ms}$ temporal resolution).  The color bar labeled \textit{Po} denotes the expert with the highest activation at each timestep chosen by the router network, illustrating the temporal specialization and switching behavior of the mixture-of-experts policy. The color bar labeled \textit{Hu} represents a human-annotated task phase segmentation of the same rollout.}
    \label{fig:full_task_expert_selection}
\end{figure*}

We evaluate LAR-MoE through a simulation benchmark and on-hardware robotic experiments. 
For quantitative assessment, we report performance on LIBERO~\cite{liu2023libero} and additionally conduct on-hardware experiments to assess real-world generalization in a surgical manipulation task. Our evaluation addresses four aspects: \textit{(i)} how do the individual components of LAR-MoE contribute to performance, \textit{(ii)} how does LAR-MoE compare to established imitation learning methods on LIBERO~\cite{liu2023libero}, \textit{(iii)} how effectively can LAR-MoE handle complex, long-horizon on-hardware tasks requiring coordinated execution of multiple subtasks, and \textit{(iv)} does the expert routing exhibit structured or interpretable patterns emerging from unsupervised training?

\subsection{How do the components of LAR-MoE contribute to performance improvements?}

\begin{figure}[t]
    \centering
    \begin{subfigure}[t]{\columnwidth}
        \centering
        \includegraphics[width=0.75\linewidth, clip, trim={17cm, 13cm, 17cm, 0cm}]{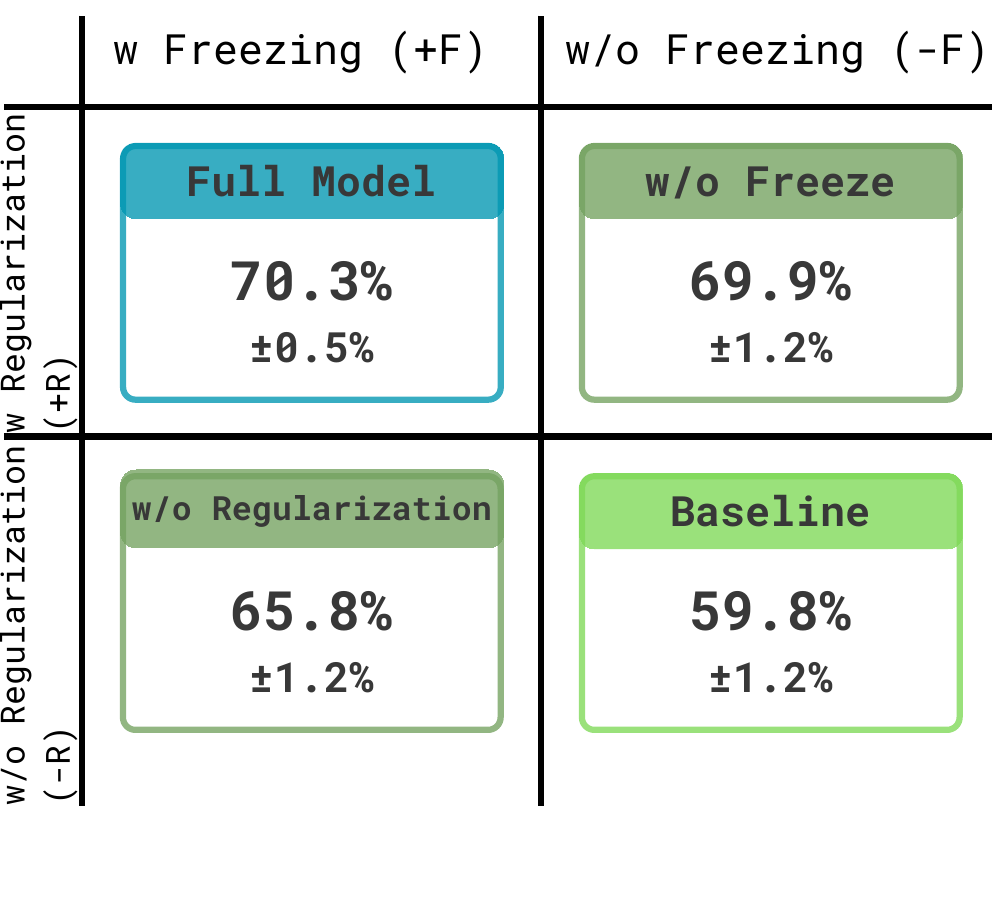}
        \caption{Router training ablation. $\pm$F: freeze student model; $\pm$R: apply proposed regularization.}
        \label{fig:design_choices}
    \end{subfigure}
    \vspace{0.3cm}
    \begin{subfigure}[t]{\columnwidth}
        \centering
        \includegraphics[width=0.75\linewidth, clip, trim={0.5cm, 0cm, 4cm, 0cm}]{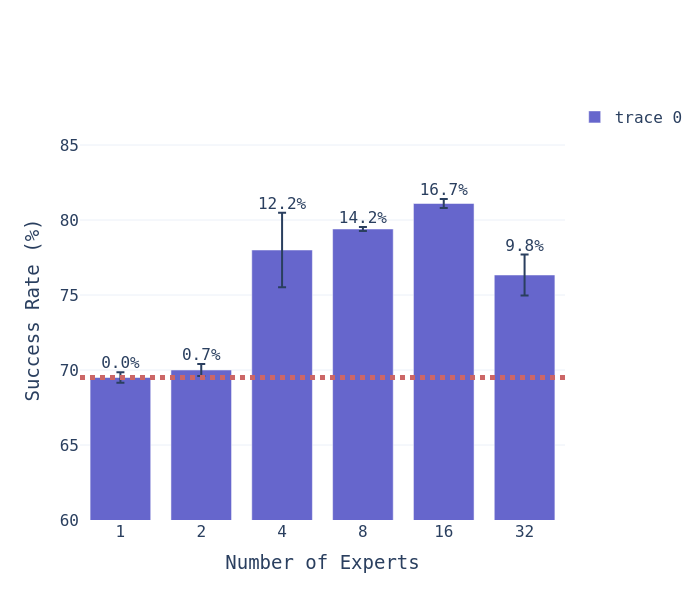}
        \caption{Impact of number of experts on LAR-MoE performance. All variants trained for 20 epochs.}
        \label{fig:number_of_experts}
    \end{subfigure}
    \caption{Ablation studies on the LIBERO benchmark~\cite{liu2023libero}. (a) Effect of freezing the student encoder (+F) and applying latent-alignment regularization (+R) on LAR-MoE16 success rate. (b) Success rate as a function of expert count.}
    \label{fig:ablations}
\end{figure}
We conduct these ablation studies on LIBERO~\cite{liu2023libero}, as it is widely adopted in the imitation learning community and provides a controlled setting to determine optimal model configuration.

\textbf{Expert Routing} We consider the naive Soft MoE variant without freezing the student encoder (\textit{-F}) and without regularization terms (\textit{-R}) as our baseline. 
Starting from this configuration, we observe a consistent improvement in success rate as we incrementally introduce freezing of the student model during expert training and the proposed latent-alignment regularization, as seen in Fig. \ref{fig:design_choices}.
Each additional design component yields a measurable gain in performance.

Despite being trained for the same number of epochs, the fully configured LAR-MoE with 16 experts achieves the highest training efficiency, reaching an improvement of $16.4\%$ over the baseline without requiring additional optimization steps. 

\textbf{Number of Experts}
Using the full configuration (+F +R), we train LAR-MoE for a total of 20 epochs with varying numbers of experts to assess how the architecture scales with additional capacity. As seen in Fig.~\ref{fig:number_of_experts}, LAR-MoE exhibits consistent performance gains as the number of experts increases up to 16, with an improvement of up to $16.7\%$ compared to the single-expert baseline, suggesting that the routing mechanism effectively leverages additional expert capacity rather than suffering from collapse or under-utilization. However, we observe a performance degradation with 32 experts, which may be attributed to insufficient training epochs for the larger model to converge.

Given that LAR-MoE with 16 experts achieves the best performance within the same training epochs, we adopt it with frozen student encoder and latent-alignment regularization as the default configuration for all subsequent experiments, as it offers a favorable trade-off between task performance and computational training cost. We refer to this model hereafter as LAR-MoE16.

\subsection{How does LAR-MoE compare to other IL methods on LIBERO?}

We train for a maximum of 100 epochs on the provided human demonstrations of the benchmark and evaluate under the standard protocol, reporting the average success rate over three runs with different random seeds. As shown in Table ~\ref{tab:libero_results}, despite its compact size of only $150$M parameters, our model outperforms several VLA models with substantially larger parameter counts and approaches the performance of $\pi_{0.5}$, which contains approximately $20\times$ more parameters.

\setlength{\tabcolsep}{2pt}
\begin{table}[h]
\centering
\caption{Success Rate (\%) on LIBERO}
\label{tab:libero_results}
\begin{tabular}{l c c c c c c}

\toprule
\textbf{Policy} 
& \textbf{Size}  
& \textbf{Spatial} 
& \textbf{Object} 
& \textbf{Goal} 
& \textbf{10} 
& \textbf{Avg.} \\
\midrule

Diffusion Policy \cite{chi2024diffusionpolicy}
& 263M
& 78.3 & 92.5 & 68.3 & 50.5 & 72.4 \\

Octo \cite{octo_2023}
& 90M
& 78.9 & 85.7 & 84.6 & 51.1 & 75.1 \\

OpenVLA \cite{kim2024openvlaopensourcevisionlanguageactionmodel}
& 8B 
& 84.7 & 88.4 & 79.2 & 53.7 & 76.5 \\

$\pi_0$ \cite{shukor2025smolvlavisionlanguageactionmodelaffordable}
& 3.5B 
& 90.0 & 86.0 & 95.0 & 73.0 & 86.0 \\

$\pi_{0.5}$ \cite{intelligence2025pi05visionlanguageactionmodelopenworld}
& 3.5B
& \textbf{98.0} & 98.0 & \textbf{98.0} & \textbf{92.0} & \textbf{97.0} \\

SmolVLA \cite{shukor2025smolvlavisionlanguageactionmodelaffordable} 
& 240M 
& 87.0 & 93.0 & 88.0 & 63.0 & 82.7 \\

SmolVLA \cite{shukor2025smolvlavisionlanguageactionmodelaffordable}
& 2.25B  
& 93.0 & 94.0 & 91.0 & 77.0 & 88.7 \\

\midrule
LAR-MoE16 (Ours) 
& \textit{150M}
& \textbf{98.0} & \textbf{99.0} & 96.0 & 88.0 & 95.2 \\
\bottomrule
\end{tabular}
\end{table}

\subsection{How effectively can LAR-MoE handle complex, long-horizon on-hardware tasks?}

\begin{figure*}[h]
    \centering
    \includegraphics[width=0.8\textwidth, trim={0cm 0cm 0cm 0cm}, clip]{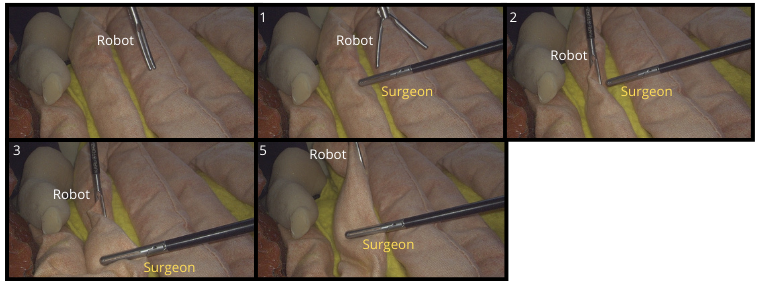}
    \caption{Illustration of the surgical bowel grasping and retraction task. The first frame shows the robot instrument, awaiting the start of the task. The remaining figures correspond to individual phases, numbered in the upper-left corner. The robot-controlled instrument is labeled in white (\textit{Robot}) and the surgeon-operated instrument in yellow (\textit{Surgeon}). Phase~1: the surgeon indicates the target grasping region. Phase~2: the robot grasps the indicated bowel segment. Phase~3: the robot holds its position while the surgeon grasps the opposite end. Phase~4 is not shown, as it represents the transitional stretching motion between Phase~3 and Phase~5. Phase~5: the robot maintains tension throughout the remainder of the procedure.}
    \label{fig:fig_phases}
\end{figure*}

We evaluate our approach on the surgical bowel grasping and retraction task introduced by~\cite{mazza2026moeact}. This task is complex, as it consists of multiple interdependent subtasks that depend on surgeon's interactions with the surgeon's instrument, requiring sustained attention and adaptive behavior from the policy. The procedure encompasses five distinct phases, and works as follows: (\textit{i}) the surgeon indicates a target region of the bowel that the policy must grasp, (\textit{ii}) the policy grasps the aforementioned region, (\textit{iii}) the policy waits without moving until the surgeon grasps the other end of the bowel, (\textit{iv}) the policy must then lift and stretch the bowel accordingly and (\textit{v}) maintain tension throughout the surgical intervention. An illustration of each phase is shown in Fig.~\ref{fig:fig_phases}.

We train our policy on the dataset collected by~\cite{mazza2026moeact} containing a total of 120 demonstrations. The observation space consists of the left endoscope image, and the action space is defined by the delta instrument tip position in Cartesian space along with a binary gripper state (open or closed). 

To calculate the success rate, two trained medical students and one surgical resident evaluate the rollout in a single-blinded process and label it as a success or failure. The final decision of the rollout is determined by majority voting. 

\begin{table}[H]
\centering
\caption{On-hardware success rates for the bowel grasping and retraction task on a bowel phantom.}
\label{table_success_rate_surgical}
\small 
\setlength{\tabcolsep}{4pt} 
\begin{tabular}{@{}lcccc@{}} 
\toprule
\multirow{2}{*}{Policy} & \multicolumn{3}{c}{Phase Success} & \multirow{2}{*}{E2E} \\
\cmidrule(lr){2-4}
& Reach & Grasp & Retract & \\
\midrule
ACT & 16/20 & 13/20 & 12/20 & 10/20 \\
ACT + Supervised MoE & 20/20 & 20/20 & 19/20 & \textbf{17/20} \\
\midrule
LAR-MoE16 (Ours) & 20/20  & 18/20  & 18/20  & \textbf{17/20}  \\
\bottomrule
\end{tabular}
\end{table}

We report the success rate of this task in table \ref{table_success_rate_surgical}, benchmarking against the supervised MoE implementation of~\cite{mazza2026moeact}. Our model successfully learns the bowel grasping and retraction task from only 120 training episodes, without requiring explicit phase annotations as used in the supervised MoE baseline. This substantially reduces labeling requirements, as surgical phase annotations are costly and rarely available, thereby enabling the use of larger unannotated data corpora. Despite relying on less labeled supervision, our MoE achieves a success rate comparable to the previously reported supervised model. Notably, one of the unsuccessful roll-outs was caused not by policy failure but by tissue slippage due to insufficient grasping force of the mechatronic interface holding the instrument.

\textbf{Zero-shot transfer to \textit{ex vivo} tissue.} To assess generalization beyond the phantom setup, we evaluate our policy zero-shot on \textit{ex vivo} porcine bowel without any additional training. Over 20 roll-outs, we achieve a success rate of \textbf{45\% (9/20)}, demonstrating the model's ability to transfer to real tissue with different visual appearance and mechanical properties. For reference, the supervised MoE baseline~\cite{mazza2026moeact} was evaluated over fewer rollouts, limiting direct statistical comparison.

\begin{figure*}[h]
    \centering
    \begin{subfigure}[t]{0.49\textwidth}
        \centering
        \includegraphics[width=\linewidth, trim={2cm 0cm 4cm 1cm}, clip]{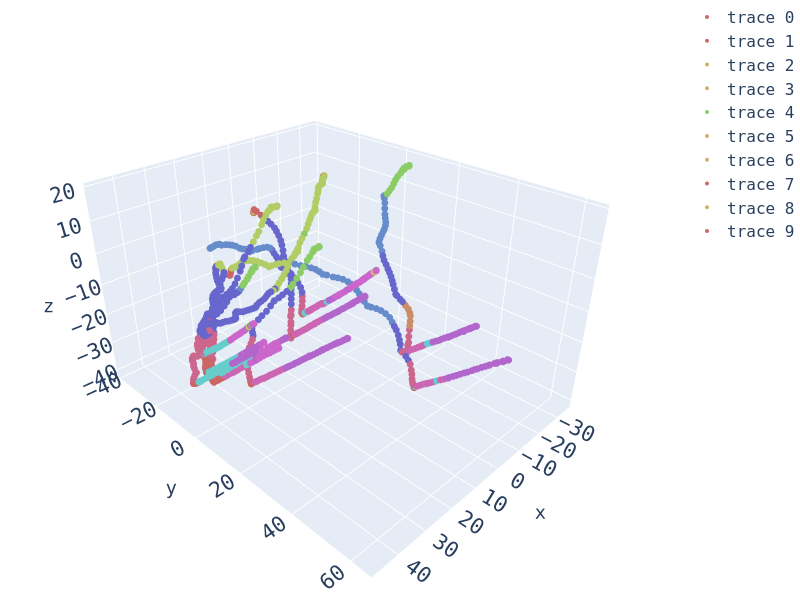}
        \caption{Phantom rollouts.}
        \label{fig:spatial_expert_phantom}
    \end{subfigure}
    \hfill
    \begin{subfigure}[t]{0.49\textwidth}
        \centering
        \includegraphics[width=\linewidth, trim={2cm 0cm 4cm 1cm}, clip]{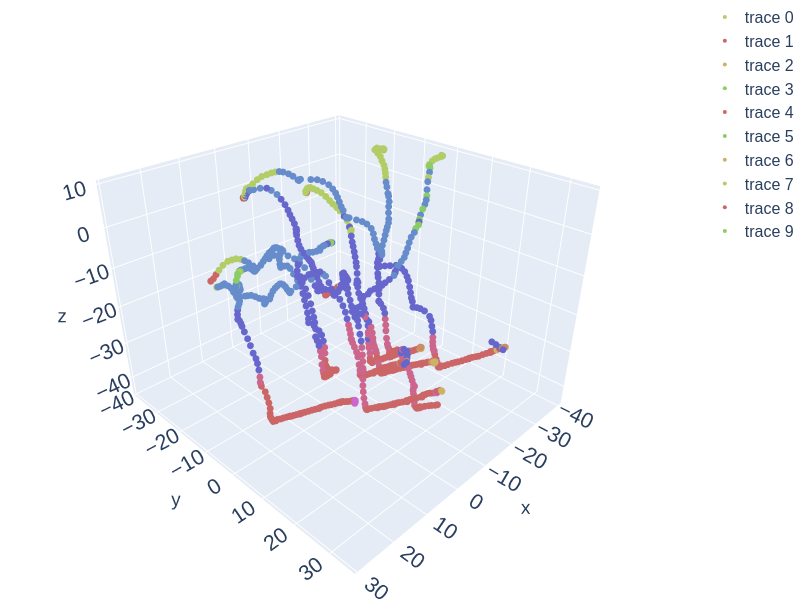}
        \caption{Ex vivo rollouts.}
        \label{fig:spatial_expert_exvivo}
    \end{subfigure}
    \caption{Spatial expert assignment during 10 rollouts on (a) phantom and (b) \textit{ex vivo} porcine tissue. Each color corresponds to a different expert and indicates the most active expert at the respective spatial position in \textit{millimeters}. Expert colors are shared across both visualizations to enable direct comparison.}
    \label{fig:spatial_expert_selection}
\end{figure*}
\subsection{Does the expert selection exhibit a structured pattern emerging from unsupervised training?}
\textbf{Unsupervised task phase partitioning} Figure~\ref{fig:full_task_expert_selection} illustrates a randomly selected \textit{ex vivo} rollout of our unsupervised MoE policy together with the corresponding expert activations over time. Although the network was not trained with explicit task phase annotations, it was able to implicitly infer meaningful phase structures solely from visual observations. The resulting partitioning into task phases generalizes to the \textit{ex vivo} setting, as evidenced by the strong similarity to a surgeon’s manual phase segmentation of the same rollout. This highlights the potential of phase-agnostic imitation learning for robust generalization.

Furthermore, the expert activation patterns are spatially consistent, as seen in Fig.~\ref{fig:spatial_expert_selection}, demonstrating that individual experts specialize in distinct regions of the task space. While the dominant experts during the initial approach and grasping phases differ between phantom and \textit{ex vivo} settings, likely due to visual domain differences, the retraction and holding phases are largely shared across both domains. This suggests that the policy learns transferable skill representations for the contact-rich manipulation phases, which may contribute to the observed zero-shot generalization performance.

\section{CONCLUSIONS}

We present LAR-MoE, a two-stage imitation learning framework that decomposes complex manipulation behaviors into specialized experts without requiring manual skill annotations. In the first stage, a student–teacher co-training strategy learns a joint latent representation that aligns visual observations with future action trajectories: a teacher encoder receives both images and action chunks while a student encoder receives only images, discovering a latent space that captures the underlying skill structure of the demonstrations without explicit supervision. In the second stage, LAR-MoE anchors soft expert routing to distances in this learned latent space, rather than relying on externally defined task partitions or supervised phase labels. This latent-aligned routing mitigates expert collapse, improves parameter efficiency, and encourages coherent specialization across heterogeneous manipulation behaviors. Our analysis on the bowel grasping and retraction task corroborates this: the learned routing reveals that experts specialize in distinct manipulation skills, producing temporal and spatial activation patterns that closely mirror human-annotated phase segmentations without ever being trained on them.

Despite its size of only 150M parameters, LAR-MoE achieves 95.2\% average success rate on the LIBERO benchmark, outperforming large-scale models with over 1B parameters and approaching state-of-the-art results.

In future work, such unsupervised task phase partitioning is not limited to scaled-up variants of the MoE architecture for imitation learning. Rather, the implicit learning of task phases from visual and kinematic data can be leveraged across a broad range of applications.

\printbibliography

\end{document}